\documentclass{llncs}
\usepackage{graphicx}
\usepackage{amsmath,amssymb} 
\usepackage{color}
\usepackage{tabularx}
\usepackage{multirow}
\usepackage{adjustbox}
\begin{document}
	\mainmatter
	
	\def\PSIVT19SubNumber{17}  
	
	\title{Enhanced Transfer Learning with ImageNet Trained Classification Layer} 
	\titlerunning{PSIVT-19 submission ID \PSIVT19SubNumber}
	\authorrunning{PSIVT-19 submission ID \PSIVT19SubNumber}
	
	\author{Tasfia Shermin\inst{1}\and
		Shyh Wei Teng\inst{1}\and
		Manzur Murshed\inst{1}\and
		Guojun Lu\inst{1}\and
		Ferdous Sohel\inst{2}\and
		Manoranjan Paul\inst{3}}
	%
	%
	\institute{School of Science, Engineering and Information Technology\\Federation University, Australia\\ \email{\{t.shermin, shyh.wei.teng, manzur.murshed, guojun.lu\}@federation.edu.au}\and
		Murdoch University, Australia\\
		\email{F.Sohel@murdoch.edu.au}\\
		\and
		Charles Sturt University, Australia\\
		\email{mpaul@csu.edu.au}}
	\maketitle              
	%
	
	%
	%
	
	%
	%
	\begin{abstract}
		Parameter fine tuning is a transfer learning approach whereby learned parameters from pre-trained source network are transferred to the target network followed by fine-tuning. Prior research has shown that this approach is capable of improving task performance. However, the impact of the ImageNet pre-trained classification layer in parameter fine-tuning is mostly unexplored in the literature. In this paper, we propose a fine-tuning approach with the pre-trained classification layer. We employ layer-wise fine-tuning to determine which layers should be frozen for optimal performance. Our empirical analysis demonstrates that the proposed fine-tuning performs better than traditional fine-tuning. This finding indicates that the pre-trained classification layer holds less category-specific or more global information than believed earlier. Thus, we hypothesize that the presence of this layer is crucial for growing network depth to adapt better to a new task. Our study manifests that careful normalization and scaling are essential for creating harmony between the pre-trained and new layers for target domain adaptation. We evaluate the proposed depth augmented networks for fine-tuning on several challenging benchmark datasets and show that they can achieve higher classification accuracy than contemporary transfer learning approaches.
		
	\end{abstract}
	\section{Introduction}
	Convolutional neural networks \cite{krizhevsky2012imagenet}, \cite{simonyan2014very}, \cite{szegedy2015going}, \cite{he2016deep} require a huge amount of labelled training data to yield optimal performance. Luckily, training CNNs on a large and diverse dataset (e.g. ImageNet) has been shown to enable the knowledge transfer across a wide range of tasks \cite{sharif2014cnn}. Parameter fine-tuning is one of the best performing transfer learning approaches used by the deep learning community. Parameter fine-tuning assists transferring learned knowledge to accomplish the target task with limited labelled data and increases the performance of the target model over random initialization \cite{yosinski2014transferable}. The sequence of traditional parameter fine-tuning is to replace the classification layer of a CNN, pre-trained on a large and diverse dataset (e.g. ImageNet), with a randomly initialized new classification layer as per the target task. Then the new model undergoes forward-backward propagation to tune gradient descent on the target set. This transfer learning approach is exploited successfully by a number of contemporary transfer learning research \cite{hariharan2015hypercolumns}, \cite{yang2015multi}, \cite{oquab2014learning}, \cite{sermanet2013overfeat}. 
	\begin{figure}[h]
		\centering
		\includegraphics[width=6cm,height=6cm,keepaspectratio]{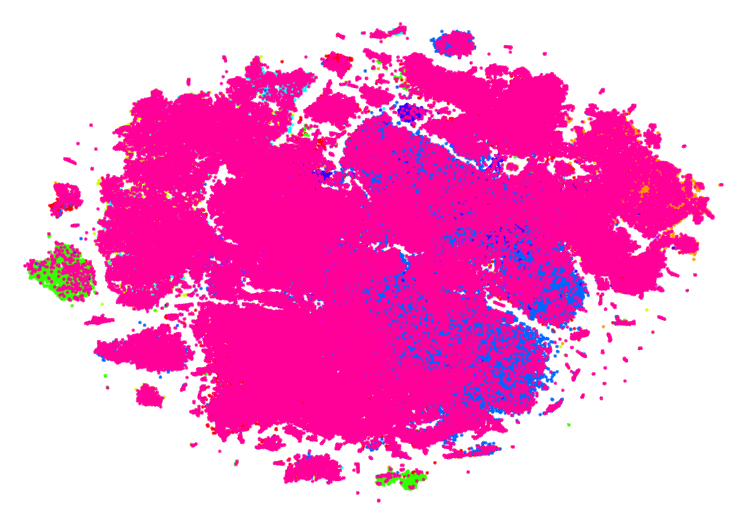}
		\caption{The t-SNE visualization of extracted features from the pre-trained classification layer. Dark pink colour represents ImageNet features, and other colours represent features from eight different target sets.}
		\label{tsne_fc8}
	\end{figure}
	
	The intuition behind so far not using the pre-trained classification layer while fine-tuning is that this layer holds category-specific features \cite{yosinski2014transferable} that may not generalize well to target sets. However, our research manifests that even the classification layers of CNNs pre-trained on ImageNet have of overlapping or neighbouring high-level features with the target sets of images of natural and artificial objects, since ImageNet consists of a massive amount of labelled images of natural and human-made objects. Fig. \ref{tsne_fc8} shows the t-SNE \cite{maaten2008visualizing} visualization of the relative distribution of extracted features from the classification layer of pre-trained AlexNet for widely used eight transfer learning target datasets (Section 4.1) and ImageNet (source). The t-SNE algorithm tries to minimize the divergence between two distribution by preserving the close or related clusters of high dimension in converted low dimension: one distribution is that measures pair-wise similarity in higher dimensional input data (in our case thousand dimensional features of source-target datasets) and the other distribution that measures pair-wise similarity in lower dimension (in our case two-dimensional space for visualization). We have used t-SNE for visualizing the relation or closeness between source and target features. The high intermingling or neighbouring between source and target feature distribution manifests that the classification layer of ImageNet pre-trained CNN may well assist the target network to adapt to the target domain via parameter fine-tuning. Also, by jointly adapting pre-trained classification and other representation layers for the target task, we could essentially bridge the domain shift underlying both the marginal distribution and the conditional distribution, which is pivotal for enhancing transfer learning \cite{zhang2013domain}. Thus, we argue that the pre-trained classification layer is important for transfer learning and propose to include this layer in the fine-tuning procedure. In this work, we evaluate all fine-tuning approaches with our layer-wise fine-tuning scheme to observe fine-tuning from which layer produces optimal performance.
	
	A significant number of works have experimented with incremental and lifelong learning \cite{sigaud2016towards}, \cite{tessler2017deep}, \cite{pickett2016growing}. For developmental transfer learning, in consistent with the traditional fine-tuning sequence, Wang et al. \cite{wang2017growing} and Oquab et al. \cite{oquab2014learning} have discarded the pre-trained classification layer and appended new layers after the penultimate fully-connected (FC) layer of AlexNet. However, inspired by our empirical analysis of the transferability of the pre-trained classification layer (Section 4.3), we argue that the presence of the pre-trained classification layer is vital to increasing the network depth for transfer learning. Thus, we propose to consider this layer as the last FC layer and to append new layers beyond it for developmental transfer learning. 
	
	During fine-tuning, our proposed depth augmented networks might struggle from internal covariate shift of activations across pre-trained and new layers. Thus, to establish harmony between the learning of new and pre-trained layers, and to reduce sensitivity to random initialization, we introduce a normalization scheme to the network. We experiment with $L_2$-norm normalization \cite{wang2017growing} and batch normalization \cite{Ioffe:2015:BNA:3045118.3045167} to search for the best performing normalization scheme for the proposed depth augmented networks.  
	\begin{flushleft}
		\textbf{The main contributions of the paper are as follows:}
	\end{flushleft}
	\begin{itemize}
		\item We propose to include the pre-trained classification layer in fine-tuning and find that the transfer learning performance with the pre-trained classification layer is higher than the traditional fine-tuning approach without it.
		\item We investigate which layers should be frozen during fine-tuning for optimal performance.
		\item For developmental transfer learning, we propose to augment new layers beyond the pre-trained classification layer to adapt better to target task. We also investigate the best fit normalization scheme for our proposed depth augmented networks.
	\end{itemize}
	The rest of the paper is organized as follows. Our proposed approaches are presented in Section 2. Section 3 presents the setup and analysis of our experimentation. Section 4 summarizes the contributions of this paper.
	
	\begin{figure}[!ht]
		\centering
		\includegraphics[width=\textwidth,height=14cm,keepaspectratio]{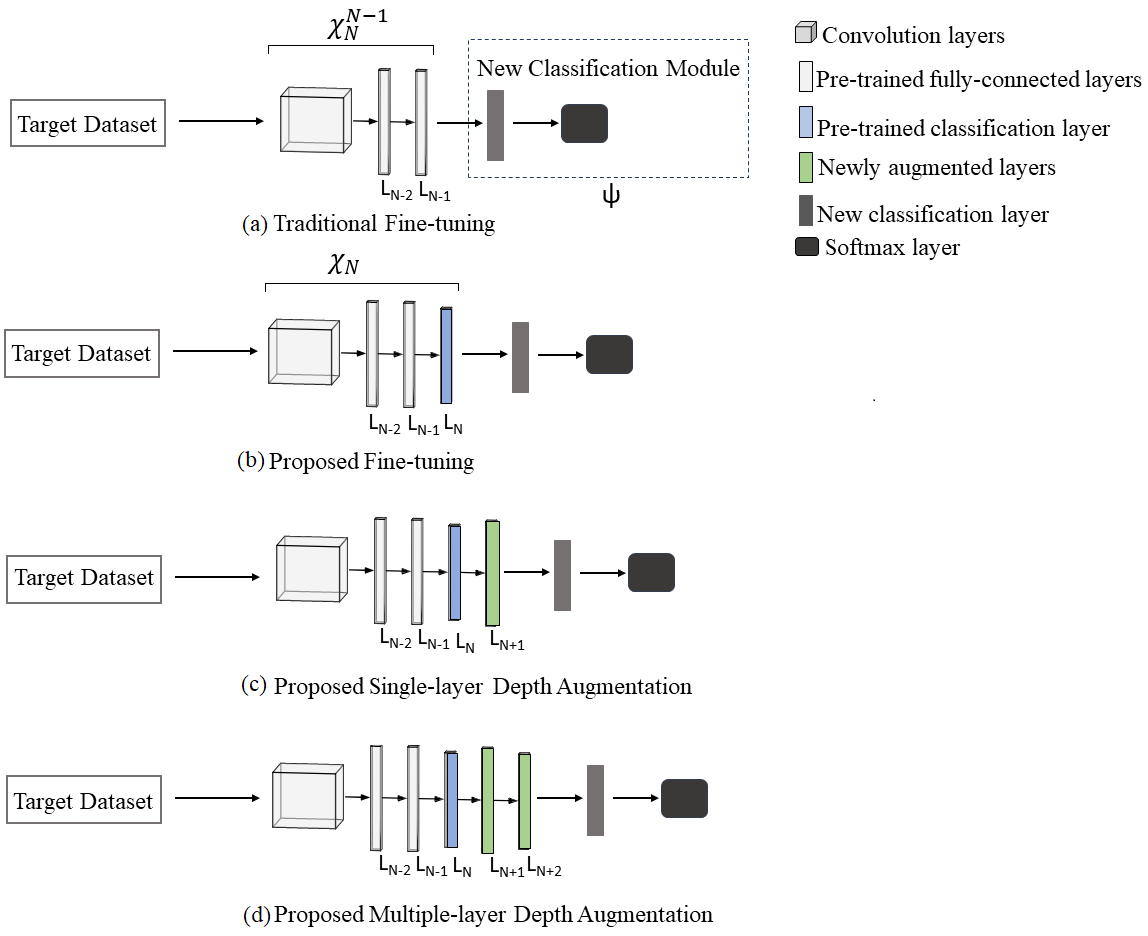}
		\caption{Block diagram of the (a) traditional fine-tuning, (b) proposed fine-tuning and (c)-(d) depth augmented networks for fine-tuning, illustrating architectures of fine-tuning TLN $[\chi_N^\kappa]_\nu^\psi$, $[\chi_N]_\nu^\psi$,  $[\chi_N]_\nu^{1+\psi}$, and $[\chi_N]_\nu^{2+\psi}$, respectively.}
		\label{grow}
	\end{figure}
	\section{Methodology}
	We introduce the architectural and notational details of the proposed and traditional fine-tuning, the proposed depth augmented networks, and the layer-wise fine-tuning scheme in this section.
	Let us assume $\chi_N$ be a CNN pre-trained with a large dataset (e.g. ImageNet) having $N$ layers $L_1, ..., L_N$, including the classification layer.
	\subsection{Traditional and Proposed Fine-tuning}
	Let $\chi_{N}^\kappa$ denote the sub-network comprising the first $\kappa$ layers of $\chi_N$, $1\leq\kappa<N$. Let $[\chi_{N}^\kappa+\psi]_\nu\equiv[\chi_{N}^\kappa ]_{\nu}^\psi$ denote a transfer-learning network (TLN) from the first $\kappa$, $1\leq\kappa<N$, layers of the pre-trained CNN $\chi_N$, with parameter fine-tuning from layer $L_\nu$ onwards, $1\leq\nu\leq\kappa$, where $\psi$ is a classification module for new classes, which is a FC classification layer $C$ followed by a Softmax layer. Fig. \ref{grow}(a) illustrates the block diagram of a TLN ($[\chi_N^{N-1}]_{\nu}^\psi$) which follows traditional fine-tuning sequence where the pre-trained classification layer $L_N$ is discarded before fine-tuning. On the contrary, we include the pre-trained classification layer $L_N$ in the proposed fine-tuning approach, as shown in Fig. \ref{grow} (b). Our proposed fine-tuning TLN which comprises of all the layers of $\chi_N$, is denoted as $[\chi_N]_{\nu}^\psi$, with parameter fine-tuning from layer $L_\nu$ onwards.
	\subsection{Proposed Depth Augmented Networks for Fine-tuning}
	We increase the depth capacity of the network by constructing new FC layers comprising $S \in \{512, 1024, 2048, 4096\}$ neurons on top of the classification layer $L_N$ as shown in Fig. \ref{grow} (c) and (d). Let $[\chi_N+L_{N+1}+...+L_{N+\tau}+\psi]_\nu\equiv[\chi_N ]_\nu^{\tau+\psi}$ denote a depth augmented TLN from the pre-trained CNN $\chi_N$ augmented with $\tau$, $\tau\geq0$, additional FC layers $L_{N+1}, ..., L_{N+\tau}$, with parameter fine-tuning from layer $L_\nu$ onwards, $1\leq\nu\leq N+\tau$, where $\psi$ is the new classification module, which has a FC classification layer $C$ with a Softmax layer. Appended layers are treated as adaptation layers to compensate for the different image statistics of the source and target sets. Moreover, they allow for suitable compositions of pre-existing parameters and avoid unwanted modifications to the parameters of pre-trained layers for their adaptation to the new task. 
	
	To maintain learning pace, we propose to include a normalization scheme in proposed depth augmented networks. We explore both $L_2$-norm normalization and batch normalization. For the first normalization approach, consistent with \cite{wang2017growing}, we apply $L_2$-norm normalization to the input activations of new layers. In case of batch normalization, we standardize the mean and variance of the input activations of new layers for stabilizing the learning process. Finally, we employ the learnable scaling parameter to scale the normalized activations.
	
	\subsection{Layer-wise Fine-tuning Scheme} 
	
	We evaluate all the approaches discussed above (Sections 2.1 and 2.2) using a two-step layer-wise fine-tuning scheme. At the first step, we initialize the transferred layers with pre-trained parameters and new layers randomly. In the second step, we start fine-tuning from the last transferred layer and freeze other layers. These two steps are repeated $K$ times with different setups ($K$ is the number of transferred layers), i.e. each time we unfreeze one more penultimate layer. For instance, in the second setup, we start fine-tuning from the penultimate transferred layer onwards. It is worth mentioning that the fine-tuning setups are mutually exclusive and parameters of all the different setups are initialized according to Step 1. We record transfer learning performance for each of these fine-tuning setups to determine which setup yields optimal performance.  
	
	\section{Performance Study and Analysis}
	This section describes the datasets used in our experiments, the implementation details, and our evaluation outcomes for proposed approaches.
	\begin{table}[!h]
		\caption{Selected datasets for target task.}
		\begin{center}
			\begin{tabular}{c|c|c|c}
				\hline
				Type & Name & Images & Categories \\ \hline\hline
				\multirow{4}{*}{Fine grained} & 102 Flowers & 8189 & 102 \\
				& CUB 200-2011 & 11788 & 200 \\ 
				&Stanford Dogs &20580&120\\
				&Oxford Pets &7400&37\\\hline
				\multirow{4}{*}{Coarse} & Caltech-256 & 30607 & 256 \\
				& Pascal VOC-07 & 9963 & 20 \\ 
				&MIT-67 scenes &15620 &67\\
				&SUN-397 scenes&108754&397\\
				\hline
			\end{tabular}
			\label{tab:T1}
		\end{center}
	\end{table}
	\subsection{Datasets and Implementation Details}
	We assembled eight different fine-grained and coarse target datasets, as stated in Table \ref{tab:T1}. Fine-grained datasets used in this work are 102 Flowers \cite{nilsback2008automated} with 102 categories, CUB 200-2011 \cite{wah2011caltech} with 200 types of birds, Stanford Dogs \cite{KhoslaYaoJayadevaprakashFeiFei_FGVC2011} with 120 classes and Oxford Pets \cite{parkhi12a} with 37 classes. The Coarse or mixed semantic datasets are Caltech-256 \cite{griffin2007caltech} with 256 categories, Pascal VOC-07 \cite{pascal-voc-2007} having 20 different classes, MIT-67 scenes \cite{quattoni2009recognizing} with 67 classes of indoor scenes and SUN-397 scenes \cite{xiao2010sun} with 397 categories. We have used ImageNet as the source dataset. 
	
	We have used AlexNet \cite{krizhevsky2012imagenet} and VGG-16 \cite{simonyan2014very} pre-trained on ImageNet as source networks. Networks with two different depths, i.e. AlexNet, and VGG-16 are used to observe whether proposed parameter fine-tuning has consistent performance across different architectures. For pre-processing the training dataset, input images are first randomly cropped, horizontally flipped, and then normalized. A split of 75\% of target dataset is used for training, and the remaining 25\% for testing. We execute 2000 iterations with a batch size of 100 and momentum 0.9 for fine-tuning. A global learning rate of 0.005 is used with a piece-wise scheduler which lowers down the learning rate by 10 times less than the previous one at every 10 epochs. We have used 10 times higher learning rate in the newly appended layers of proposed depth augmented networks.

	\begin{figure}[!h]
		\centering
		\includegraphics[width=\textwidth,height=8cm,keepaspectratio]{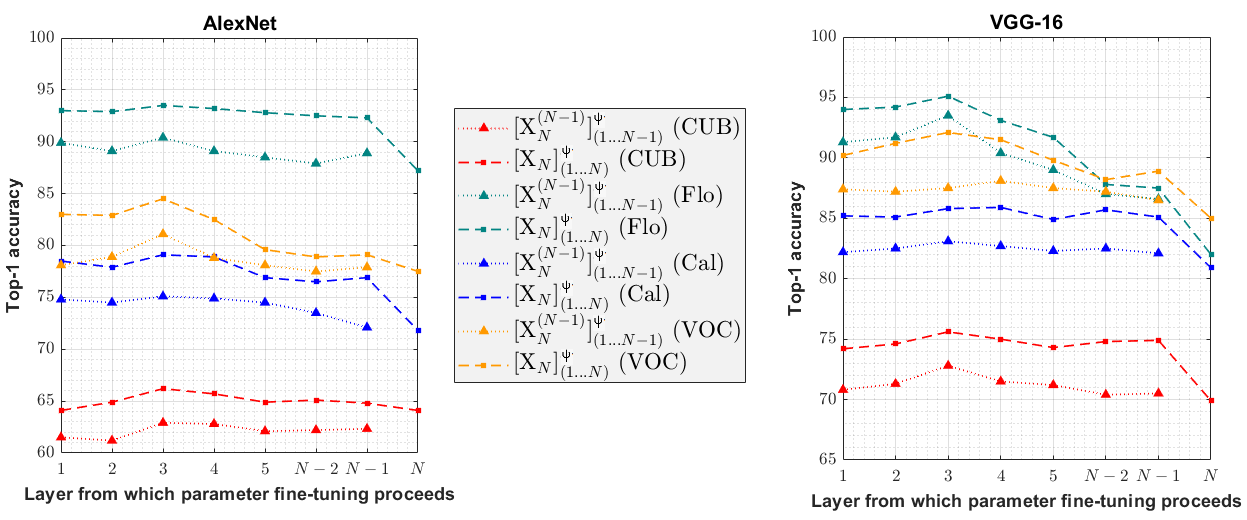}
		\caption{Both graphs present the performance of proposed fine-tuning with the pre-trained classification layer (dashed lines), and traditional fine-tuning (dotted lines) for CUB 200-2011, 102 Flowers, Caltech-256 and Pascal VOC-07 datasets. (a) Here, the \textit{x-axis} represents 8 layers of the AlexNet as $1, 2,..., N-2, N-1, N$ and shows the layer from which fine-tuning proceeds while earlier layers are frozen (e.g. $N-2$ denotes fine-tuning of layer $L_{N-2}$ to $L_N$ and other layers are frozen). (b) Here, the \textit{x-axis} shows 16 layers (5 convolution blocks and 3 FC layers) of the VGG-16 as $1, 2, ..., N-2, N-1, N$. For both networks, proposed fine-tuning significantly outperforms traditional fine-tuning for all datasets.}
		\label{av1}
	\end{figure}
	\subsection{Evaluation and Analysis of Proposed Fine-tuning}
	To investigate the impact of pre-trained classification layer in parameter fine-tuning and to compare the performance of proposed fine-tuning with traditional fine-tuning, we utilize our layer-wise fine-tuning scheme (Section 2.3). Fig.~\ref{av1} presents the results of two coarse (Caltech-256 and Pascal VOC-07) and two fine-grained datasets (CUB 200-2011, 102 Flowers). Traditional fine-tuning setup (dotted lines) lags far behind the proposed fine-tuning (dashed lines) consistently for almost all fine-tuning setups. This finding substantiates that fine-tuning pre-trained classification layer along with other transferred layers assist better transfer learning. Note that other datasets also perform similarly. 
	\subsection{Layers to be Frozen for Optimal Performance}
	Proposed and traditional fine-tuning performance seems to increase when we continue to unfreeze and fine-tune more pre-trained layers according to different layer-wise fine-tuning setups (Section 2.3), as shown in Fig. \ref{av1} (a) and Fig. \ref{av1} (b). However, we observe a significant drop in performance when we tune initial convolutional layers, more specifically, convolutional layer 1 and 2 for AlexNet, and convolution block 1 and 2 for VGG-16. The intuition is that fine-tuning these generic layers might introduce noisy or unwanted modifications of parameters. That is, updating parameters would force the network to learn highly generic features of target set which are already learned from source set. The fine-tuning procedure has far fewer data and iterations than training from scratch, which might not let a vast number of pre-trained parameters of initial convolutional layers find another such equilibrium to interact with next convolutional layer in the same pace. Fig. \ref{av1}(a) portrays that fine-tuning from the third convolution layer onwards of AlexNet yields the highest accuracy. Fig. \ref{av1}(b) manifests that VGG-16 holds a similar trend for the third convolution block, which gives another perception that the first two convolution blocks of VGG-16 may contain highly generic or low-level features.
	
	\begin{figure}
		\centering
		\includegraphics[width=\textwidth,height=8cm,keepaspectratio]{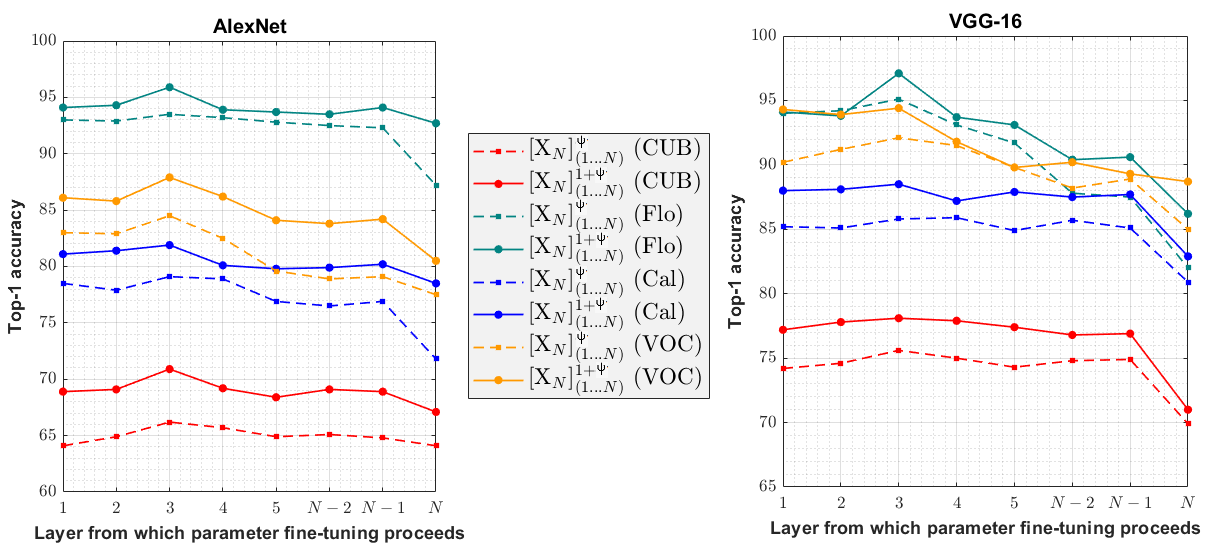}
		\caption{These accuracy graphs present the performance of proposed fine-tuning with the pre-trained classification layer (dashed lines), and proposed single-layer depth augmented network (solid lines) for CUB 200-2011, 102 Flowers, Caltech-256 and Pascal VOC-07 datasets. (a) Here, the \textit{x-axis} represents 8 layers of the AlexNet as $1, 2,..., N-2, N-1, N$ and shows the layer from which fine-tuning proceeds while earlier layers are frozen (e.g. $N-2$ denotes fine-tuning of layer $L_{N-2}$ to $L_N$ and other layers are frozen). (b) Here, the \textit{x-axis} shows 16 layers (5 convolution blocks and 3 FC layers) of the VGG-16 as $1, 2, ..., N-2, N-1, N$. Single-layer depth augmented networks further boost performance over proposed fine-tuning for all datasets.}
		\label{av2}
	\end{figure}
	\begin{table}[!h]
		\caption{Diagnostic performance analysis of our depth augmented networks. Here, $S$ denotes the number of neurons in newly appended FC layer $L_{N+1}$. Fine-tuning from layer $N-5$ (i.e. third convolution layer/ block) yields best performance for almost all combinations.}
		\begin{center}
			\begin{tabular}{c|c|c|c|c|c|c|c}
				\hline
				Network & Dataset & $S$ & $[\chi_N]_N^{(1+\psi)}$ & $[\chi_N]_{N-1}^{(1+\psi)}$ & $[\chi_N]_{N-2}^{(1+\psi)}$ &$[\chi_N]_{N-5}^{(1+\psi)}$ & All \\ \hline\hline
				&\multirow{2}{*}{CUB 200-2011} &512 & 65.2 & 66.5& 66.9 & 67.8 & 67.2 \\
				&& 1024 & 66.2 & 67.5& 68.9& 69.9& 68.4 \\ 
				&& 2048 & 66.3 & 68.7 & 69.0 & \textbf{70.9}& 69.2\\
				&& 4096 & 66.7 & 68.1 & 67.1 & 68.1& 66.8\\\cline{2-8}
				AlexNet&\multirow{2}{*}{Caltech-256} & 512 & 75.5 & 78.2& 78.6 & 79.5 & 79.1 \\
				&&1024 & 75.9 & 78.3 & 78.9 & 80.9& 80.3\\
				&&2048 & 75.2 & 77.9 & 78.5 & \textbf{81.9} & 81.1\\
				&& 4096 & 75.0 & 78.1 & 78.3 & 81.0 & 80.9\\
				\hline
				&\multirow{2}{*}{CUB 200-2011} &512 & 76.4 & 76.8 & 77.1 & 77.9 & 77.6\\
				&& 1024 & 76.8 & 77.1 & 77.5 & 77.6 & 77.9\\
				&& 2048 & 76.7 & 77.6 & 77.9 & \textbf{78.1} & 77.1\\
				&& 4096 & 75.9 & 77.7 & 77.0 & 77.5 & 77.4\\\cline{2-8}
				VGG-16&\multirow{2}{*}{Caltech-256} & 512 & 85.5 & 86.1 & 87.7 & 87.9 & 87.1 \\
				&& 1024 & 85.9 & 86.3 & 86.9 & 87.8& 87.5\\
				&& 2048 & 85.5 & 87.8 & 88.0 & 88.4 & 88.1\\
				&& 4096 & 85.3 & 88.1 & 89.5 & \textbf{88.5} & 88.1\\
				\hline
			\end{tabular}
			
		\end{center}
		
		\label{diag_grow}
	\end{table}
	\subsection{Performance Analysis of Proposed Depth Augmented Networks} 
	We append a new FC layer on top of the pre-trained classification layer, employ normalization scheme to the augmented network, and perform layer-wise fine-tuning. We discuss details about our normalization scheme later in this section. Our results shown in solid lines of Fig. \ref{av2}(a) and \ref{av2}(b) present our best performing augmented networks with 2048 neurons and signify that parameter fine-tuning with increased network capacity paves the way to learn better. Our proposed single-layer depth augmented network performed better than fine-tuning with the pre-trained classification layer. This observation verifies the effectiveness of increasing model capacity beyond the pre-trained classification layer when adapting it to both fine-grained and coarse novel classification task. Considering the best layer-wise fine-tuning performance, CUB and 102 Flowers seem to achieve more gain than the other two. 
	
	A detailed analysis of our investigation with single-layer depth augmented networks having different combinations of neurons, such as 512, 1024, 2048, and 4096, is shown in Table \ref{diag_grow}. Our empirical results indicate that the increase in performance is proportional to the increase in the magnitude of the new layer; however, for 4096 neurons, it diminishes marginally. In proposed depth augmentation approach, the bridge between new and pre-trained layers is the layer consisting only 1000 neurons; it might suffer from an overabundance of parameters while propagating information through to four times larger neural layer. Single-layer depth augmentation with 2048 neurons happens to yield best performance. It is worth mentioning that our augmented networks also perform similarly for other datasets.\\
	\begin{table}[!h]
		\caption{Performance comparison of proposed single-layer depth augmented fine-tuned AlexNet network with contemporary fine-tuning approaches.}
		\begin{center}
			\begin{adjustbox}{max width=\textwidth}
				\begin{tabular}{c|c|c|c|c|c|c|c|c}
					\hline
					Approach&\textbf{CUB 200-2011} &\textbf{102 Flowers}&\textbf{Stanford Dogs}&\textbf{Oxford Pets}&\textbf{Caltech-256}&\textbf{VOC07}&\textbf{MIT-67}&\textbf{SUN-397} \\ \hline\hline
					Normal\_FT\_CNN & 62.3 &  88.9& 63.8& 80.0 & 72.1 &77.9& 61.2&53.9\\ 
					CNN-SVM \cite{sharif2014cnn} & 53.3 &74.7  &66.8  & 79.6& 72.3 & 75.3 &58.4&55.9\\
					CNNAug-SVM \cite{sharif2014cnn} & 61.8 & 86.8 & 66.6 & 79.9 & 74.8 & 76.8 & 69.0&56.2\\
					LSVM \cite{qian2015fine} & 61.4 &  87.1 &65.0 & 77.6& 69.7 & 75.2 &66.7&55.8\\
					MsML+ \cite{qian2015fine}& 66.6 & 89.4 &69.5& 81.1& 68.4 &74.8 &59.8 &52.1\\
					CombinedAlexNet \cite{joulin2016learning} &63.3 &83.3& 64.5 &76.9 &69.2 &77.1 &58.8&54.2\\ 
					WA-CNN \cite{wang2017growing}& 69.0 &92.8 & 66.9 & 82.4& 79.5 & 83.4&66.3&58.3\\
					Grow-conv \cite{azizpour2015generic} & 66.1 & 91.4 & 67.2 &  82.1& 78.3 & 77.0 &70.1 &50.2\\
					Proposed network & \textbf{70.9} & \textbf{95.9} &68.7 & \textbf{84.5}& \textbf{81.9} & \textbf{87.9} &69.9 &\textbf{62.6}\\
					\hline
				\end{tabular}%
			\end{adjustbox}
		\end{center}
		
		\label{alex}
	\end{table}
	\begin{table}[!h]
		\caption{Performance comparison of proposed single-layer depth augmented fine-tuned VGG-16 network with contemporary fine-tuning approaches.}
		\begin{center}
			\begin{adjustbox}{max width=\textwidth}
				\begin{tabular}{c|c|c|c|c|c|c|c|c}
					\hline
					Approach&\textbf{CUB 200-2011} &\textbf{102 Flowers}&\textbf{Stanford Dogs}&\textbf{Oxford Pets}&\textbf{Caltech-256}&\textbf{VOC07}&\textbf{MIT-67}&\textbf{SUN-397} \\ \hline\hline
					Normal\_FT\_CNN & 70.5 & 85.6 & 68.2 & 85.2 & 83.9 &86.5&66.5 &61.8\\ 
					CNN-SVM \cite{sharif2014cnn} & 66.5 &81.5 & 66.7 &86.4  & 79.9 & 82.4 &60.4&56.6\\
					Muldip-Net \cite{tamaazousti2017learning} & 71.5 & 81.9 & 65.0 & 86.1 & 80.9 &87.5  &68.9 &63.5\\
					Grow-conv \cite{azizpour2015generic} & 72.5 &  88.7& 75.1 & 89.1 & 86.1 & 89.1&72.1 &67.5\\
					DA-CNN \cite{wang2017growing} & 76.1 &93.3  & 72.8&88.4& 84.9 &91.4 &73.1 &67.2\\
					Proposed network & \textbf{78.1} &\textbf{97.1}  &\textbf{76.1}  &\textbf{90.6}  & \textbf{88.5} &\textbf{94.4} &\textbf{76.8} &\textbf{69.8}\\
					\hline
				\end{tabular}
			\end{adjustbox}
		\end{center}
		
		\label{vgg}
	\end{table}\\
	\textbf{Comparison with Contemporary Transfer Learning Works} To further prove the robustness of proposed single-layer depth augmented networks, we summarize the performance comparison with different existing transfer learning approaches from literature in Tables \ref{alex} and \ref{vgg}. The best outcomes among various combinations of our single-layer depth augmented AlexNet and VGG-16 evaluated by our layer-wise fine-tuning scheme are shown. For other approaches, the performance gap between our implementation and that reported by \cite{sharif2014cnn}, \cite{qian2015fine}, \cite{joulin2016learning}, \cite{azizpour2015generic}, \cite{wang2017growing}, \cite{tamaazousti2017learning} is due to different target sets, train-test splits, network architectures, and iterations. Note that we have used similar hyper-parameters, iterations, and train-test splits for all approaches in Tables \ref{alex} and \ref{vgg} to maintain a fair comparison. Consistent superior outcomes validate that the presence of the pre-trained classification layer in increasing model capacity for parameter fine-tuning is effective for adjusting the network to a wide range of target tasks.
	\begin{table}[h]
		\caption{Performance comparison between single-layer and multiple-layer depth augmented networks. Only the highest performing combination is presented here. Multiple-layer depth augmented networks for other datasets also yield consistent performance.}
		\begin{center}
			\begin{tabular}{c|c|c|c}
				\hline
				Network & Dataset & Configuration & Accuracy (\%) \\  \hline\hline
				& \multirow{2}{*}{CUB 200-2011} & $[\chi_N]_{N-5}^{1+\psi}$ & 70.9\\
				& & $[\chi_N]_{N-5}^{2+\psi}$ & \textbf{72.1}\\\cline{2-4}
				& \multirow{2}{*}{102 Flowers} & $[\chi_N]_{N-5}^{1+\psi}$  & 95.9\\
				& & $[\chi_N]_{N-5}^{2+\psi}$ & \textbf{97.2}\\\cline{2-4}
				AlexNet	&  \multirow{2}{*}{Caltech-256} & $[\chi_N]_{N-5}^{1+\psi}$ & 81.9\\ 
				& & $[\chi_N]_{N-5}^{2+\psi}$ & \textbf{82.8} \\\cline{2-4}
				&  \multirow{2}{*}{VOC-07} & $[\chi_N]_{N-5}^{1+\psi}$ & 87.9\\
				& & $[\chi_N]_{N-5}^{2+\psi}$ & \textbf{88.8} \\
				\hline
				& \multirow{2}{*}{CUB 200-2011} &$[\chi_N]_{N-5}^{1+\psi}$ & 78.1\\
				& & $[\chi_N]_{N-5}^{2+\psi}$ & \textbf{78.9}\\\cline{2-4}
				& \multirow{2}{*}{102 Flowers} &$[\chi_N]_{N-5}^{1+\psi}$  & 97.1\\
				& & $[\chi_N]_{N-5}^{2+\psi}$& \textbf{98.2}\\\cline{2-4}
				VGG-16&  \multirow{2}{*}{Caltech-256} & $[\chi_N]_{N-5}^{1+\psi}$ & 88.5\\ 
				& & $[\chi_N]_{N-5}^{2+\psi}$ & \textbf{89.4}\\\cline{2-4}
				&  \multirow{2}{*}{VOC-07} & $[\chi_N]_{N-5}^{1+\psi}$  & 94.4\\
				& &$[\chi_N]_{N-5}^{2+\psi}$ & \textbf{95.9} \\
				\hline
			\end{tabular}
			
		\end{center}
		
		\label{multi_grow}
	\end{table}
	\begin{table}
		\caption{Performance comparison between single-layer depth augmented networks with $L_2$-norm and batch-norm normalization scheme.}
		\begin{center}
			\begin{tabular}{c|c|c|ccccc}
				\hline
				\multirow{2}{*}{Network} & \multirow{2}{*}{Dataset} & \multirow{2}{*}{Norm}&\multicolumn{5}{c}{Accuracy (\%)}\\  \cline{4-8}
				& & & $[\chi_N]_N^{(1+\psi)}$ & $[\chi_N]_{N-1}^{(1+\psi)}$ & $[\chi_N]_{N-2}^{(1+\psi)}$& $[\chi_N]_{N-5}^{(1+\psi)}$ & All\\
				\hline\hline
				\multirow{2}{*}{AlexNet} & \multirow{2}{*}{CUB 200-2011} & Standardization & 66.2 & 67.5 & 68.9 & \textbf{69.9} & 68.4\\
				& & $L_2$ & 62.1 & 64.2 & 64.3 & \textbf{64.5} & 63.5\\
				\hline
				\multirow{2}{*}{VGG-16} & \multirow{2}{*}{CUB 200-2011} & Standardization & 76.7 & 77.6 & 77.9 & \textbf{78.1} & 78.5\\
				& & $L_2$ & 71.5 & 72.1 & 73.1 & \textbf{73.2}& 72.7\\
				\hline
			\end{tabular}
		\end{center}
		
		\label{BN}
	\end{table}\\
	\textbf{Comparison of Single and Multiple-layer Depth Augmentation} Augmenting two new layers beyond pre-trained classification layer is observed to be the cut-off point as performance starts to diminish after that. Table \ref{multi_grow} shows the results of the best combination (i.e. $L_{N+1}$=2048 and $L_{N+2}$=1024) of two-layer and single-layer depth augmentation. Appending two new layers after the pre-trained classification layer facilitate network marginally over single-layer augmentation by increasing representational capacity. It is proven once again that the pre-trained classification layer holds prominent high-level features which are capable of propagating learned knowledge to multiple newly appended layers. This also manifests increasing network incrementally by augmenting depth is a stable parameterization for improving performance.
	
	\subsubsection{Best Fit Normalization Scheme} After exploring two types of normalization, we observe that standardization \cite{Ioffe:2015:BNA:3045118.3045167} assisted better learning for proposed single and multiple-layer depth augmented networks. We represent the results of diagnostic experiments with $S = 1024$ and standardization in Table \ref{BN}. Results of the single-layer depth augmented AlexNet trained on CUB 200-2011 dataset show that the improvement of our depth augmented network is around 2\% compared to traditional fine-tuning for $L_2$-norm normalization, and more than 6\% otherwise. A similar significant boost in performance is also noticed in other datasets, which are not stated in this paper for limited space. Increase in task performance states that standardization reduces the chances of the pre-trained activations to dominate the randomly initialized ones. 
	\begin{table}[]
		\caption{Performance gain of proposed fine-tuning $[\chi_{N}]_{N-5}^\psi$ from traditional fine-tuning $[\chi_{N}^{N-1}]_{N-5}^\psi$, single-layer depth augmented networks $[\chi_{N}]_{N-5}^{1+\psi}$ from proposed fine-tuning, and double-layer $[\chi_{N}]_{N-5}^{2+\psi}$ from single-layer depth augmented networks respectively for fine-grained datasets, where parameter fine-tuning proceeds from layer $L_{N-5}$.}
		\begin{center}
			\resizebox{\textwidth}{!}{%
				\begin{tabular}{c|c|c|c|c|c|c|c|c|c|c}
					\hline
					& \multicolumn{2}{|c|}{CUB 200-2011}& \multicolumn{2}{|c|}{102 Flowers}& \multicolumn{2}{|c|}{Stan. Dogs}&\multicolumn{2}{|c|}{Oxf. Pets}& \multicolumn{2}{|c}{Avg.}\\\hline
					& AlexNet & VGG-16 & AlexNet & VGG-16 & AlexNet & VGG-16 & AlexNet & VGG-16 & AlexNet & VGG-16\\\hline\hline
					$[\chi_{N}]_{N-5}^\psi$ &3.3&2.8&3.1&1.6&1.4&1.6&4.2&2.0&\textbf{3.0}&\textbf{2.0}\\\hline
					$[\chi_{N}]_{N-5}^{1+\psi}$ &4.7&2.5&2.0&2.9&1.8&3.7&2.1&3.7&\textbf{2.7}&\textbf{3.2}\\\hline
					$[\chi_{N}]_{N-5}^{2+\psi}$ &1.2&0.8&1.3&1.1&0.4&1.2&1.0&0.9&\textbf{1.0}&\textbf{1.0}\\\hline
				\end{tabular}
			}
		\end{center}
		
		\label{l3}
	\end{table}
	\begin{table}[]
		\caption{Performance gain of proposed fine-tuning $[\chi_{N}]_{N-5}^\psi$ from traditional fine-tuning $[\chi_{N}^{N-1}]_{N-5}^\psi$, single-layer depth augmented networks $[\chi_{N}]_{N-5}^{1+\psi}$ from proposed fine-tuning, and double-layer $[\chi_{N}]_{N-5}^{2+\psi}$ from single-layer depth augmented networks respectively for coarse datasets, where parameter fine-tuning proceeds from layer $L_{N-5}$.}
		\begin{center}
			\resizebox{\textwidth}{!}{%
				\begin{tabular}{c|c|c|c|c|c|c|c|c|c|c}
					\hline
					& \multicolumn{2}{c|}{Caltech-256}&\multicolumn{2}{|c|}{VOC-07}&\multicolumn{2}{|c|}{MIT-67}&\multicolumn{2}{|c|}{SUN-397}&  \multicolumn{2}{|c}{Avg.}\\\hline
					& AlexNet & VGG-16 & AlexNet & VGG-16 & AlexNet & VGG-16 & AlexNet & VGG-16 & AlexNet & VGG-16 \\\hline\hline
					$[\chi_{N}]_{N-5}^\psi$ &4.0&2.7&3.4&4.6&5.1&6.5&6.7&4.5&\textbf{4.8}&\textbf{4.6}\\\hline
					$[\chi_{N}]_{N-5}^{1+\psi}$ &2.8&2.7&3.4&2.3&3.6&3.8&2.0&3.5&\textbf{3.0}&\textbf{3.1}\\\hline
					$[\chi_{N}]_{N-5}^{2+\psi}$ &0.9&0.9&0.9&1.5&1.4&1.1&1.7&1.3&\textbf{1.2}&\textbf{1.2}\\\hline
				\end{tabular}
			}
		\end{center}
		
		\label{l4}
	\end{table}
	\subsection{Average Performance Gain} Tables \ref{l3} and \ref{l4} show that in average both coarse and fine-grained target datasets leverage the presence of the pre-trained classification layer in proposed fine-tuning and depth augmentation. However, coarse datasets manifest slightly more significant performance gain than fine-grained ones. This suggests that this layer possesses general information to transfer to a wide range of datasets. Results show that both the networks gain significant improvement for single-layer augmentation while for two-layer, the increase is marginal. Moreover, less deep backbone network seems to be more benefited from depth augmentation.
	
	\section{Conclusion}
	In this paper, we demonstrate that the ImageNet and widely used transfer learning target sets have neighbouring high-level features, therefore, adapting the pre-trained classification layer which catches high-level features would help fine-tuning. We propose a novel fine-tuning approach with the pre-trained classification layer. We empirically establish that proposed fine-tuning approach outperforms traditional fine-tuning for all selected target datasets. Also, we notice on average, the coarse target datasets with ImageNet achieve more performance gain than fine-grained ones. For evaluating traditional and proposed fine-tuning approaches, we use our layer-wise fine-tuning scheme. Our layer-wise fine-tuning scheme manifests that freezing initial convolutional layers yield optimal fine-tuning performance for all target datasets. 
	
	Being inspired by developmental transfer learning and impact of the pre-trained classification layer in fine-tuning, we augment new layers beyond the pre-trained classification layer for a better adaptation of the target task. Moreover, to tune pre-trained and new parameters at a steady speed and to encourage better learning, we normalize and scale input activations of augmented layers. Assessment of the proposed depth augmented networks on eight different datasets show that they outperform existing transfer learning approaches. Our empirical study has provided practitioners strong justification to utilize the ImageNet pre-trained classification layer for fine-tuning and depth augmentation beyond it for adapting the network to target tasks.
	%
	%
	

\end{document}